\documentclass[article]{jss}

\usepackage{thumbpdf}
\usepackage{amssymb} 
\usepackage{subcaption} 
\usepackage{amsmath} 
\usepackage{array} 
\usepackage{mleftright}
\usepackage{xparse}
\usepackage{framed}
\usepackage{float}

\newcommand{\class}[1]{`\code{#1}'}
\newcommand{\fct}[1]{\code{#1()}}

\NewDocumentCommand{\evalat}{sO{\big}mm}{%
  \IfBooleanTF{#1}
   {\mleft. #3 \mright|_{#4}}
   {#3#2|_{#4}}%
}

\author{Jaime Pizarroso\\Universidad Pontificia\\ Comillas
   \And Jos\'e Portela\\Universidad Pontificia\\ Comillas
   \And Antonio Mu\~noz\\Universidad Pontificia\\ Comillas}
\Plainauthor{Jaime Pizarroso, Jos\'e Portela, Antonio Mu\~noz}

\title{\pkg{NeuralSens}: Sensitivity Analysis of Neural Networks}
\Plaintitle{NeuralSens: Sensitivity Analysis of Neural Networks}
\Shorttitle{NeuralSens: Sensitivity Analysis of Neural Networks}

\Abstract{
This article presents the \pkg{NeuralSens} package that can be used to perform sensitivity analysis of neural networks using the partial derivatives method. The main function of the package calculates the partial derivatives of the output with regard to the input variables of a multi-layer perceptron model, which can be used to evaluate variable importance based on sensitivity measures and characterize relationships between input and output variables. Methods to calculate partial derivatives are provided for objects trained using common neural network packages in R, and a \class{numeric} method is provided for objects from packages which are not included.  The package also includes functions to plot the information obtained from the sensitivity analysis. 

The article contains an overview of techniques for obtaining information from neural network models, a theoretical foundation of how partial derivatives are calculated, a description of the package functions, and applied examples to compare \pkg{NeuralSens} functions with analogous functions from other available \proglang{R} packages.
}

\Plainkeywords{neural networks, sensitivity analysis, variable importance, R, neuralsens}

\Address{
  Jaime Pizarroso, Jos\'e Portela, Antonio Mu\~noz  \\
  Instituto de Investigación Tecnológica (IIT) \\
  Escuela Técnica Superior de Ingeniería ICAI \\
  Universidad Pontificia Comillas \\
  Calle de Alberto Aguilera, 23\\
  28015 Madrid, Spain\\
  E-mail: \email{Jaime.Pizarroso@iit.comillas.edu},\\
  \email{jose.portela@iit.comillas.edu}, \\
  \email{Antonio.Munoz@iit.comillas.edu}
}

\begin{document}

\section[Introduction]{Introduction} \label{sec:intro}

As the volume of available information increases in various fields, the number of situations where data-intensive analysis can be applied also grows simultaneously (\cite{philip_chen_data_intensive_2014},\cite{valduriez_scientific_2018}).This analysis can be used to extract useful information and supports decision-making (\cite{sun_big_2018}). 

Machine-learning algorithms are commonly used in data-intensive analysis (\cite{hastie01statisticallearning}, \cite{butler_machine_2018}, \cite{vu_shared_2018}), as they are able to detect patterns and relations in the data without being explicitly programmed. Artificial Neural Networks (ANN) are one of the most popular machine-learning algorithms due to their versatility. ANNs were designed to mimic the biological neural structures of animal brains (\cite{mcculloch_logical_nodate}) by ``learning'' to perform tasks by considering examples and modifying their structure through iterative algorithms (\cite{rojas_fast_1996}). The form of ANN that is discussed in this paper is the feed-forward multilayer perceptron (MLP) (\cite{rumelhart_learning_1986}). MLPs are one of the most popular form of ANNs and have been used in a wide variety of applications (\cite{mosavi_state_2019}, \cite{smalley_ai-powered_2017}, \cite{hornik_multilayer_1989}). This model consists of interconnected units, called nodes or perceptrons, that are arranged in layers. The first layer consists of inputs (or independent variables), the final layer is the output layer, and the layers in between are known as hidden layers (\cite{ozesmi_artificial_1999}). Assuming that there is a relationship between the outputs and the inputs, the goal of the MLP is to approximate a non-linear function to represent the relationship between the output and the input variables of a given dataset with minimal residual error (\cite{hornik_approximation_nodate}, \cite{cybenko_approximation_1989}).

Neural networks provide predictive advantages when compared to other models, such as the ability to implicitly detect complex non-linear relationships between dependent and independent variables. However, the complexity of neural networks makes it difficult to obtain information on how the model uses the input variables to predict the output. Finding methods for extracting information on how the input variables affect the response variable has been a recurrent topic of research in neural networks (\cite{olden_accurate_2004}, \cite{zhang_opening_2018}). Some examples are:
\begin{enumerate}
	\item Neural Interpretation Diagram (NID) as described by \cite{ozesmi_artificial_1999} for plotting the ANN structure. A NID is a modified version of the standard representation of neural networks which changes the color and thickness of the connections between neurons based on the sign and magnitude of its weight. 
	\item Garson's method for variable importance \citep{garson_interpreting_1991}. It consists of summing the product of the absolute value of the weights connecting the input variable to the response variable through the hidden layer. Afterwards, the result is scaled relative to all other input variables. The relative importance of each input variable is given as a value from zero to one.
	\item Olden's method for variable importance \citep{olden_accurate_2004}. This method is similar to Garson's, but it uses the real value instead of the absolute value of the connection weights and it does not scale the result.
	\item Input Perturbation (\cite{scardi_developing_1999}, \cite{gevrey_review_2003}). It consists of adding an amount of white noise to each input variable while maintaining the other inputs at a constant value. The resulting change in a chosen error metric for each input perturbation represents the relative importance of each input variable.
	\item Profile method for sensitivity analysis (\cite{lek_application_1996}). Similar to the Input Perturbation algorithm, it changes the value of one input variable while maintaining the other variables at a constant value. These constant values are different quantiles of each variable, therefore a plot of model predictions across the range of values of the input is obtained. A modification of this algorithm is proposed in \cite{beck_neuralnettools_2018}. To avoid unlikely combinations of input values, a clustering technique is applied to the training dataset and the center values of the clusters are used instead of the quantile values.
	\item Partial derivatives method for sensitivity analysis (\cite{dimopoulos_use_1995}, \cite{dimopoulos_neural_1999}, \cite{muoz_variable_1998}, \cite{white_statistical_2001}). It performs a sensitivity analysis by computing the partial derivatives of the ANN outputs with regard to the input neurons evaluated on the samples of the training dataset (or an analogous dataset).
	\item Partial dependence plot (PDP) (\cite{friedman_greedy_2001}, \cite{ICE}). PDPs help visualize the relationship between a subset of the input variables and the response while accounting for the average effect of the other inputs. For each input, the partial dependence of the response with regard to the selected input is calculated following two steps. Firstly, individual conditional expectation (ICE) curves are obtained, one for each sample of the training dataset. The ICE curve for sample $k$ is built by obtaining the model response using the input values at sample $k$, except for the input variable of interest, whose value is replaced by other values it has taken in the training dataset. Finally, the PDP curve for the selected variable is calculated as the mean of the ICE curves obtained.
	\item Local interpretable model-agnostic explanations (\cite{ribeiro_why_2016}). The complex neural network model is explained by approximating it locally with an interpretable model, such as a linear regression or a decision tree model.
	\item Forward stepwise addition (\cite{gevrey_review_2003}). It consists of rebuilding the neural network by sequentially adding an input neuron and its corresponding weights. The change in each step in a chosen error metric represents the relative importance of the corresponding input.
	\item Backward stepwise elimination (\cite{gevrey_review_2003}). It consists of rebuilding the neural network by sequentially removing an input neuron and its corresponding weights. The change in each step in a chosen error metric represents the relative importance of the corresponding input.
\end{enumerate}

These methods help with neural network diagnosis by retrieving useful information from the model. However, these methods have some disadvantages: NID can be difficult to interpret given the amount of connections in most networks, Garson's and Olden's algorithms only account for the weights of the input variable connections in the hidden layer, and Lek's profile method may present analyses of input scenarios not represented by the input training data or require other methods like clustering (using the center of the clusters instead of the range quantiles of the input variables) with its inherent disadvantages (\cite{xu_comprehensive_2015}). Partial dependence plots have a similar disadvantage as they might provide misleading information if the value of the output variable depends not only on the variable of interest but also on compound effects of input variables. Local linearization is useful for interpreting the input variable importance in specific regions of the dataset, but it does not give a quantitative importance measure for the entire dataset.  Forward stepwise addition and backward stepwise elimination perform a more exhaustive analysis, but are computationally expensive and may produce different results based on the order in which the inputs are added/removed and the initial training conditions of each model.
 
The partial derivatives method overcomes these disadvantages by analytically calculating the derivative of each output variable with regard to each input variable evaluated on each data sample of a given dataset. The contribution of each input is calculated in both magnitude and sign taking into account not only the connection weights and the activation functions, but also the values of each input. By using all the samples of the dataset, the effect of the input variables in the response is calculated for the real values of the data, avoiding information loss due to clustering. Analytically calculating the derivatives results in more robust diagnostic information, because it depends solely on how well the neural network predicts the output. As long as the neural network predicts the output variable with enough precision, the derivatives will be the same regardless of the training conditions and the structure of the network \citep{beck_neuralnettools_2018}. 

As stated before, the main objective of the proposed methods  is to extract information from a given neural network model. For example, unnecessary inputs may lead to a higher complexity of the neural structure and prevent finding the optimal model, thus, affecting the performance of the neural network. Several researchers defend the ability of the partial derivatives method to determine whether an explanatory variable is irrelevant for predicting the response of the neural network (\cite{white_statistical_2001}, \cite{zurada_sensitivity_1994}, \cite{goos_determining_1995}). Pruning the neural network of these irrelevant inputs improves the capability of the neural network to model the relationship between response and explanatory variables and, consequently, the quality of  information that can be extracted from the model.

Using the partial derivatives method has some disadvantages that should be noted. The operations required to calculate partial derivatives are time-consuming when compared to other methods such as Garson's and Olden's. The computing time grows  as the size of the neural network or the size of the database used to calculate the partial derivatives increases. Additionally, the input variables should be normalized when using this method, as otherwise the value of the partial derivatives may depend on the scale of each variable and produce misleading results. However, its advantages with regard to other methods make sensitivity analysis a very useful technique for interpreting and improving neural network models.

This article describes the \pkg{NeuralSens} package (\cite{neuralsens}) for \proglang{R} (\cite{R-project}) which can be used to perform sensitivity analysis of MLP neural networks using partial derivatives. The main function of the package includes methods for MLP objects from the most popular neural network packages available in \proglang{R}. To the authors' knowledge, there is no other \proglang{R} package that calculates the partial derivatives of a neural network.  The \pkg{NeuralSens} package is available at the Comprehensive R Archive Network (CRAN) at \url{https://CRAN.R-project.org/package=NeuralSens}, and the development version is maintained as a GitHub repository at \url{https://github.com/JaiPizGon/NeuralSens}. It should be mentioned that other algorithms to analyze neural networks are already implemented in \proglang{R}: NID, Garson's, Olden's and Lek's profile algorithms are implemented in \pkg{NeuralNetTools} (\cite{beck_neuralnettools_2018}), the partial dependence plots method is implemented in \pkg{pdp} (\cite{pdp}) and local linearization is implemented in \pkg{lime} (\cite{lin_pedersen_understanding_2019}). 

The rest of this article is structured as follows. Section \ref{sec:theoretical foundation} describes the theory of the functions in the \pkg{NeuralSens} package, along with references to general introductions to neural networks. Section \ref{sec:package description} presents the architecture details of the package. Section \ref{sec:further examples} shows applied examples for using the \pkg{NeuralSens} package, comparing the results with packages currently available in \proglang{R}. Finally, Section \ref{sec:conclusions} concludes the article.

\section{Theoretical foundation} \label{sec:theoretical foundation}

The \pkg{NeuralSens} package has been designed to calculate the partial derivatives of the output with regard to the inputs of a MLP model in \proglang{R}. The remainder of this section explains the theory of multilayer perceptron models, how to calculate the partial derivatives of the output of this type of model with regard to its inputs and some sensitivity measures proposed by the authors.

\subsection{Multilayer perceptron}
\begin{figure}[t]
\centering
\includegraphics[width=0.6\linewidth, height=0.4\linewidth]{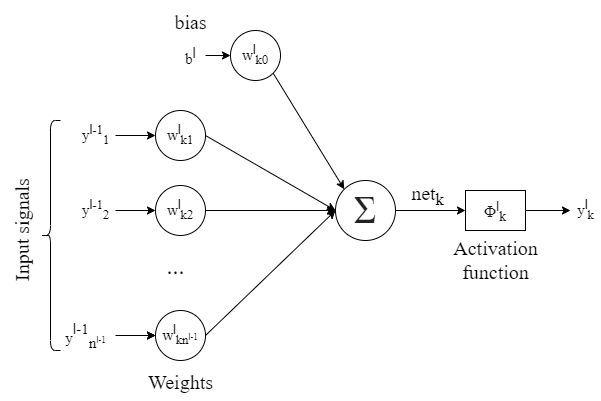}
\caption{\label{fig:neuron} Scheme of the $k^{th}$ neuron in the $l^{th}$ layer of a MLP model. $\phi^l_k$ represent the activation function of the neuron, $b^l$ represent the bias of the $l^{th}$ layer, $y^{l-1}_j$ represent the output of the $j^{th}$ neuron in the previous layer and $w^{l}_{jk}$ represent the weight of the connection between the neuron and the $j^{th}$ neuron of the previous layer.}
\end{figure}
A fully-connected feed-forward MLP has one-way connections from the units of one layer to all neurons of the subsequent layer. Each time the output of one unit travels along one connection to another unit, it is multiplied by the weight of the connection. At each unit the inputs are summed and a constant, or bias, is added. Once all the input terms of each unit are summed, an activation function is applied to the result. 

Figure \ref{fig:neuron} shows the scheme of a neuron in a MLP model and represent graphically the operations in Equation (\ref{eqn:output_l_output_l-1}).

For each neuron, the output $y^l_k$ of the $k^{th}$ neuron in the $l^{th}$ layer can be calculated by:
\begin{equation}
\label{eqn:output_l_output_l-1}
	y^l_k = \phi^l_k\left(z^l_k\right) = \phi^l_k\left(\sum_{j=1}^{n^{l-1}} w^{l}_{kj} \cdot y^{l-1}_{j} + w^l_{k0} \cdot b^l\right)
\end{equation}
where $z^l_k$ refers to the weighted sum of the neuron inputs, $n^{l-1}$ refers to the number of neurons in the $(l-1)^{th}$ layer, $w^{l}_{kj}$ refers to the weight of the connection between the $j^{th}$ neuron in the $(l-1)^{th}$ layer and the $k^{th}$ neuron in the $l^{th}$ layer, $\phi^l_k$ refers to the activation function of the $k^{th}$ neuron in $l^{th}$ layer, $b^l$ refers to the bias in the $l^{th}$ layer and $\cdot$ refers to the scalar product operation. 
For the input layer thus holds $l = 1$, $y^{1-1}_j = x_j$, $w^{1}_{kj} = 1$ and $b^1 = 0$. 

\begin{figure}[t!]
\centering
\includegraphics[width=0.85\linewidth]{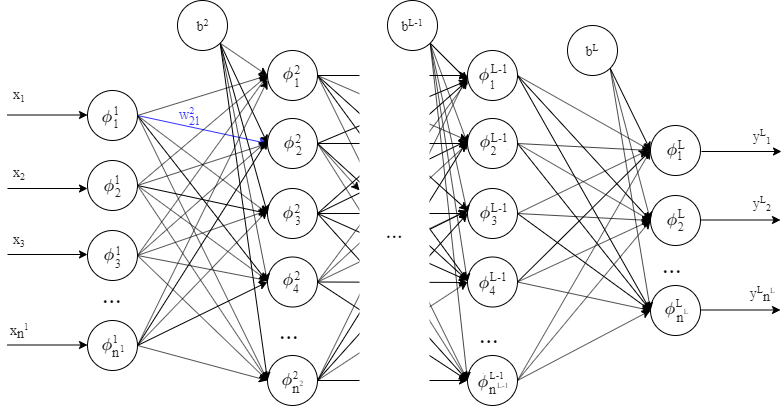}
\caption{\label{fig:mlpstruct} General MultiLayer Perceptron structure with $L$ layers. $\phi^i_j$ represent the activation function of the $j^{th}$ neuron in the $i^{th}$ layer, $b^i$ represent the bias of the $i^{th}$ layer, $x_k$ represent the input variables and $y_k$ represent the output variables.}
\end{figure}

Figure \ref{fig:mlpstruct} can be treated as a general MLP model. A MLP can have $L$ layers, and each layer $l$ ($1 \leqslant l \leqslant L)$ has $n^l$ ($n^l \geqslant 1$) neurons. $n^1$ stands for the input layer and $n^L$ for the output layer. For each layer $l$ the input dimension is equal to the output dimension of layer $(l-1)$. For a neuron $i$ ($1 \leqslant i \leqslant n^l$) in layer $l$, its input vector, weight vector and output are $\mathbf{yb}^{l-1} = \left(b^l, y^{l-1}_1, \cdots , y^{l-1}_{n^{l-1}}\right)$, $\mathbf{w}^{l}_i = \left(w^l_{i0}, w^l_{i1}, \cdots , w^l_{in^{l-1}}\right)^\top$ and $y^l_i = \phi^l_i\left(z^l_i\right) = \phi^l_i\left(\mathbf{yb}^{l-1} \cdot \mathbf{w}^{l}_i\right)$ respectively, where $\phi^l_i:\mathbb{R}\rightarrow \mathbb{R}$ refers to the neuron activation function and $\cdot$ refers to the matrix multiplication operator. For each layer $l$, its input vector is $\mathbf{yb}^{l-1}$, its weight matrix is $\mathbf{W}^{l} = \left[\mathbf{w}^l_1 \cdots \mathbf{w}^l_{n^l} \right]$ and its output vector is $\mathbf{y}^l=\left(y^l_i, \cdots , y^l_{n^l}\right) = \Phi^l\left(\mathbf{z}^l\right) = \Phi^l\left(\mathbf{yb}^{l-1} \cdot \mathbf{W}^{l}\right)$, where $\Phi^l:\mathbb{R}^{n^l} \rightarrow \mathbb{R}^{n^l}$ is a vector-valued function defined as $\Phi^l(\mathbf{z}) = (\phi^l_1(z_1), \cdots , \phi^l_{n^{l}}(z_{n^{l}}))$. 

Weights in the neural structure determine how the information flows from the input layer to the output layer. Identifying the optimal weights that minimize the prediction error of a dataset is called training the neural network. There are different algorithms to identify these weights, being the most used the backpropagation algorithm described in \cite{rumelhart_learning_1986}. Explaining these training algorithms are out of the scope of this paper.

\subsection{Partial derivatives}

The sensitivity analysis performed by the \pkg{NeuralSens} package is based on the partial derivatives method. This method consists in calculating the derivative of the output with regard to the inputs of the neural network. These partial derivatives are called sensitivity, and are defined as:
\begin{equation}
\label{eqn:sensitivity}
	\evalat[\big]{s_{ik}}{\mathbf{x}_n} = \frac{\partial y_k}{\partial x_i} \left(\mathbf{x}_n\right)
\end{equation}
where $\mathbf{x}_n$ refers to the $n$ sample of the dataset used to perform the sensitivity analysis and $\evalat[\big]{s_{ik}}{\mathbf{x}_n}$ refers to the sensitivity of the output of the $k^{th}$ neuron in the output layer with regard to the input of the $i^{th}$ neuron in the input layer evaluated in $\mathbf{x}_n$. We calculate these sensitivities applying the chain rule to the partial derivatives of the inner layers (derivatives of Equation (\ref{eqn:output_l_output_l-1}) for each neuron in the hidden layers). The partial derivatives of the inner layers are defined following the next equations:

\begin{itemize}
	\item Derivative of $z^l_k$ (input of the $k^{th}$ neuron in the $l^{th}$ layer) with regard to $y^{l-1}_i$ (output of the $i^{th}$ neuron in the $(l-1)^{th}$ layer). This partial derivative corresponds to the weight of the connection between the $k^{th}$ neuron in the $l^{th}$ layer and the   $i^{th}$ neuron in the $(l-1)^{th}$ layer:
	\begin{equation}
	\label{eqn:der_input_l_output_l-1}
		\frac{\partial z^l_k}{\partial y^{l-1}_i} = w^l_{ki}
	\end{equation}
	\item Derivative of $y^l_k$ (output of the the $k^{th}$ neuron in the $l^{th}$ layer) with regard to $z^l_i$ (input of the $i^{th}$ neuron in the $l^{th}$ layer):
	\begin{equation}
	\label{eqn:der_output_l_input_l}
		\evalat*{\frac{\partial y^l_k}{\partial z^l_i}}{z^l_i} = \frac{\partial \phi^l_k}{\partial z^l_i} \left(z^l_i\right)
	\end{equation}
	where $\frac{\partial \phi^l_k}{\partial z^l_i}$ refers to the partial derivative of the activation function of the $k^{th}$ neuron in the $l^{th}$ layer with regard to the input of the $k^{th}$ neuron in the $l^{th}$ layer evaluated for the input $z^l_i$ of the $i^{th}$ neuron in the $l^{th}$ layer.
\end{itemize}

Equations (\ref{eqn:der_input_l_output_l-1}) and (\ref{eqn:der_output_l_input_l}) have been implemented in the package in matrix form to reduce computational time following the next equations:
\begin{equation}
\label{eqn:matriz_weights_l}
	\frac{\partial \mathbf{z}^l_{[1 \times n^l]}}{\partial \mathbf{y}^{l-1}_{[1 \times n^{l-1}]}} =\left[ {\begin{array}{cccc}
	\frac{\partial z^l_1}{\partial y^{l-1}_1} & \frac{\partial z^l_2}{\partial y^{l-1}_1} & \cdots & \frac{\partial z^l_{n^l}}{\partial y^{l-1}_1} \\
	\frac{\partial z^l_1}{\partial y^{l-1}_2} & \frac{\partial z^l_2}{\partial y^{l-1}_2} & \cdots & \frac{\partial z^l_{n^l}}{\partial y^{l-1}_2} \\
	\vdots & \vdots & \ddots & \vdots \\
	\frac{\partial z^l_1}{\partial y^{l-1}_{n^{l-1}}} & \frac{\partial z^l_2}{\partial y^{l-1}_{n^{l-1}}} & \cdots & \frac{\partial z^l_{n^l}}{\partial y^{l-1}_{n^{l-1}}} \\
	\end{array} } \right]  =  \left[ {\begin{array}{cccc}
	w^l_{11} & w^l_{21} & \cdots & w^l_{n^l1} \\
	w^l_{12} & w^l_{22} & \cdots & w^l_{n^l2} \\
	\vdots & \vdots & \ddots & \vdots \\
	w^l_{1n^{l-1}} & w^l_{2n^{l-1}} & \cdots & w^l_{n^ln^{l-1}} \\
	\end{array} } \right] = \mathbf{W}^{*l}_{[n^{l-1} \times n^{l}]}
\end{equation}
\begin{equation}
\label{eqn:matriz_derivatives_l}
	\mathbf{J}^l_{l_{[n^l \times n^l]}} = \frac{\partial{\mathbf{y}^l_{[1 \times n^l]}}}{\partial{\mathbf{z}^l_{[1 \times n^l]}}} = \left[ {\begin{array}{cccc}
	\frac{\partial y^l_1}{\partial z^{l}_1} & \frac{\partial y^l_2}{\partial z^{l}_1} & \cdots & \frac{\partial y^l_{n^l}}{\partial z^{l}_1} \\
	\frac{\partial y^l_1}{\partial z^{l}_2} & \frac{\partial y^l_2}{\partial z^{l}_2} & \cdots & \frac{\partial y^l_{n^l}}{\partial z^{l}_2} \\
	\vdots & \vdots & \ddots & \vdots \\
	\frac{\partial y^l_1}{\partial z^{l}_{n^{l}}} & \frac{\partial y^l_2}{\partial z^{l}_{n^{l}}} & \cdots & \frac{\partial y^l_{n^l}}{\partial z^{l}_{n^{l}}} \\
	\end{array} } \right] = \left[ {\begin{array}{cccc}
	\frac{\partial \phi^l_1}{\partial z^{l}_1}\left(z^l_1\right) & \frac{\partial \phi^l_2}{\partial z^{l}_1}\left(z^l_1\right) & \cdots & \frac{\partial \phi^l_{n^l}}{\partial z^{l}_1}\left(z^l_1\right) \\
	\frac{\partial \phi^l_1}{\partial z^{l}_2}\left(z^l_2\right) & \frac{\partial \phi^l_2}{\partial z^{l}_2}\left(z^l_2\right) & \cdots & \frac{\partial \phi^l_{n^l}}{\partial z^{l}_2}\left(z^l_2\right) \\
	\vdots & \vdots & \ddots & \vdots \\
	\frac{\partial \phi^l_1}{\partial z^{l}_{n^{l}}}\left(z^l_{n^{l}}\right) & \frac{\partial \phi^l_2}{\partial z^{l}_{n^{l}}}\left(z^l_{n^{l}}\right) & \cdots & \frac{\partial \phi^l_{n^l}}{\partial z^{l}_{n^{l}}}\left(z^l_{n^{l}}\right) \\
	\end{array} } \right]
\end{equation}
where $\mathbf{W}^{*l}$ is the reduced weight matrix of the $l^{th}$ layer and $\mathbf{J}^l_{l}$ is the Jacobian matrix of the outputs in the $l^{th}$ layer with respect to the inputs in the $l^{th}$ layer.

Following the chain rule, the Jacobian matrix of the outputs in the $l^{th}$ layer with regard to the inputs in the $(l-j)^{th}$ layer can be calculated by:
\begin{equation}
\label{eqn:matrix_derivatives_l_l-1}
	\mathbf{J}^l_{l-j_{[n^l \times n^{j}]}} = \prod_{h = j}^{l-1} (\mathbf{J}^h_{h_{[n^h \times n^{h}]}} \cdot \mathbf{W}^{*(h + 1)}_{[n^{h} \times n^{h+1}]}) \cdot \mathbf{J}^l_{l_{[n^l \times n^{l}]}}
\end{equation}
where $1 \leqslant k \leqslant (l-1)$ and $2 \leqslant l \leqslant L$. Using this equation with $l = L$ and $j = 1$, the partial derivatives of the outputs with regard to the inputs of the MLP are obtained. 

\subsection{Sensitivity measures} \label{subsec:sens_meas}

Once the sensitivity has been obtained for each variable and observation, different measures can be calculated to analyze the results. The authors propose the following sensitivity measures to summarize the information obtained by evaluating the sensitivity of the outputs for all the input samples $X_n$ of the provided dataset:
\begin{itemize}
	\item \textit{Mean sensitivity} of the output of the $k^{th}$ neuron in the output layer with regard to the $i^{th}$ input variable:
	\begin{equation}
	\label{eqn:M_AV_S}
		S_{ik}^{avg} = \frac{\sum_{j = 1}^{N} \evalat[\big]{s_{ik}}{\mathbf{x}_j}}{N}
	\end{equation}
	where $N$ is the number of samples in the dataset. 
	\item \textit{Sensitivity standard deviation} of the output of the $k^{th}$ neuron in the output layer with regard to the $i^{th}$ input variable:
	\begin{equation}
	\label{eqn:STD_S}
		S_{ik}^{sd} = \sigma\left(\evalat[\big]{s_{ik}}{\mathbf{x}_j}\right); j \in {1,...,N}
	\end{equation}
	where $N$ is the number of samples in the dataset and $\sigma$ refers to the standard deviation function. 
	\item \textit{Mean squared sensitivity}   of the output of the $k^{th}$ neuron in the output layer with regard to the $i^{th}$ input variable (\cite{yeh_first_2010}, \cite{zurada_sensitivity_1994}):
	\begin{equation}
	\label{eqn:M_SQ_AV_S}
		S_{ik}^{sq} = \sqrt{\frac{\sum_{j = 1}^{N} \left(\evalat[\big]{s_{ik}}{\mathbf{x}_j}\right)^2}{N}}
	\end{equation}
	where $N$ is the number of samples in the dataset.
\end{itemize}

In case there are more than one output neuron, such as in a multi-class classification problem, these measures can be generalized to obtain sensitivity measures of the whole model as follows:
\begin{itemize}
	\item \textit{Mean sensitivity} with regard to the $i^{th}$ input variable:
	\begin{equation}
	\label{eqn:global_M_AV_S}
		S_{i}^{avg} = \frac{\sum_{k=1}^{n^l}S_{ik}^{avg}}{n^L}
	\end{equation}
	\item \textit{Sensitivity standard deviation} with regard to the $i^{th}$ input variable:
	\begin{equation}
	\label{eqn:global_STD_S}
		S_{i}^{sd} = \sqrt{\frac{\sum_{k=1}^{n^l}\left(\left(S_{ik}^{sd}\right)^2 + \left( S_{ik}^{avg} - S_{i}^{avg}\right)^2\right)}{n^L}}
	\end{equation}
	\item \textit{Mean squared sensitivity} with regard to the $i^{th}$ input variable \citep{yeh_first_2010}:
	\begin{equation}
	\label{eqn:global_M_SQ_AV_S}
		S_{i}^{sq} = \frac{\sum_{k=1}^{n^l} S_{ik}^{sq}}{n^L}
	\end{equation}
\end{itemize}

Methods in \pkg{NeuralSens} to calculate the sensitivities of a neural network and the proposed sensitivities measures were written for several \proglang{R} packages that can be used to create MLP neural networks: class \class{nn} from \pkg{neuralnet} package (\cite{neuralnet}), class \class{nnet} from \pkg{nnet} package (\cite{nnet}), class \class{mlp} from \pkg{RSNNS} (\cite{RSNNS}), classes \class{H2ORegressionModel} and \class{H2OMultinomialModel} from \pkg{h2o} package (\cite{h2o}), \class{list} from \pkg{neural} package (\cite{neural}) and class{nnetar} from \pkg{forecast} package (\cite{forecast}). The same methods are applied to neural network objects created with the \fct{train} function from the \pkg{caret} package (\cite{caret}) only if these \class{train} objects inherit from the available packages the ``class'' attribute. Methods have not been included in \pkg{NeuralSens} for other packages that can create MLP neural networks, although further developments of \pkg{NeuralSens} could include additional methods. An additional method for class \class{numeric} is available to use with the basic information of the model (weights, structure and activation functions of the neurons). Examples on how to use this \class{numeric} method can be found in appendix \ref{app:ext_pack_func_impl}.

\section{Package structure} \label{sec:package description}

The functionalities of the package \pkg{NeuralSens} is based on the new \proglang{R} class \class{SensMLP} defined inside the package itself. \pkg{NeuralSens} includes four main functions based on this class to perform the sensitivity analysis of a MLP model described in the previous section:
\begin{itemize}
	\item \fct{SensAnalysisMLP}: \code{S3} method to perform the sensitivity analysis using partial derivatives of the outputs with regard to the inputs of the MLP model. This function returns a \class{SensMLP} object with the results of the sensitivity analysis.
	\item \fct{SensitivityPlots}: graphically represent the sensitivity measures of a \class{SensMLP} object.
	\item \fct{SensFeaturePlot}: graphically represent the relation between the sensitivities of a \class{SensMLP} object and the value of the input variables.
	\item \fct{SensTimePlot}: graphically represent the evolution among time of the sensitivities of a \class{SensMLP} object.
\end{itemize}

Each of these functions are detailed in the rest of this section. The output of the last three functions are plots created with \pkg{ggplot2} package functions \citep{ggplot2}.

\subsection[The R class `SensMLP']{The \proglang{R} class \class{SensMLP}}\label{subsec:sensmlpclass}

The \pkg{NeuralSens} package defines an S3 object called \class{SensMLP} as a list with the following components:
\begin{itemize}
	\item sens: \class{list} of \class{data.frames}, one per neuron in the output layer, with the $S_{ik}^{avg}$, $S_{ik}^{sd}$ and $S_{ik}^{sq}$ sensitivity measures described in Section \ref{subsec:sens_meas} (Equations (\ref{eqn:M_AV_S}), (\ref{eqn:STD_S}) and (\ref{eqn:M_SQ_AV_S})). Each row of the \code{data.frame} contains the sensitivity measures with regard to a specific input.
	\item raw\_sens: \class{list} of \class{matrixes}, one per neuron in the output layer, with the sensitivities calculated following Equation (\ref{eqn:matrix_derivatives_l_l-1}) with $l = 1$ and $L = L$. Each column of each \class{matrix} contains the sensitivities of the output with regard to a specific input and each row contains the sensitivities with regard to all the inputs corresponding to the same row of the trData component.
	\item mlp\_struct: \class{numeric} \class{vector} indicating the number of neurons in each layer of the MLP model.
	\item trData: typically a \class{data.frame} which contains the dataset used to calculate the sensitivities.
	\item coefnames: \class{character} \class{vector} with the names of the input variables of the MLP model.
	\item output\_name: \class{character} \class{vector} with the names of the output variables of the MLP model.
\end{itemize}
Functions described in Sections \ref{subsec:Vis_Sensitivity} (\fct{SensitivityPlots}),  \ref{subsec:Vis_Sensitivity_time} (\fct{SensTimePlot}) and \ref{subsec:Vis_Sensitivity_inputs} (\fct{SensFeaturePlot}) can be accessed through the plot method of the \class{SensMLP} object. \fct{print} and \fct{summary} methods are also available for obtaining information on the sensitivities and sensitivity measures of the \class{SensMLP} object. Examples of these methods are presented in the remaining sections.

\subsection{MLP Sensitivity Analysis}\label{subsec:Sensitivity}

The \fct{SensAnalysisMLP} function calculates the partial derivatives of a MLP model. This function consists of an \class{S3} method \citep{s3_method} to extract the basic information of the model (weights, structure and activation functions) based on the model \code{class} attribute and to pass this information to the default method. This default method calculates the sensitivities of the model as described in Section \ref{sec:theoretical foundation}, and creates a \code{SensMLP} object with the result of the sensitivity analysis. \fct{SensAnalysisMLP} function performs all operations using matrix calculus to reduce the computational time. 

In the current version of \pkg{NeuralSens} (version 1.0.0), the accepted activation functions are shown in Table \ref{tab:act_functs}. To calculate the sensitivities, the function assumes that all the neurons in a defined layer has the same activation function.

\begin{table}[t!]
  \begin{center}
    \begin{tabular}{>{\centering\arraybackslash}l|>{\centering\arraybackslash}l>{\centering\arraybackslash}l}
     \hline
     \textbf{Name} & \textbf{Function} & \textbf{Derivative} \\
     \hline
      \code{sigmoid}
      & $f(z) = \frac{1}{1+e^{-z}}$ & $\frac{\partial f}{\partial z}(z) = \frac{1}{1+e^{-z}}\cdot \left(1-\frac{1}{1+e^{-z}}\right)$\\[10pt]
      \code{tanh}
      & $f(z) = \tanh{(z)}$ & $\frac{\partial f}{\partial z}(z) = 1 - \tanh{(z)}$\\[10pt]
      \code{linear}
      & $f(z) = z$ & $f'(z) = 1$\\[10pt]
      \code{ReLU}
      & $f(z) = \begin{cases}
				  \begin{aligned}
				  	0 & \quad \text{ when } z \leqslant 0 \\
				  	z & \quad \text{ when } z > 0 \\
				  \end{aligned}
               \end{cases}$ & 
               $\frac{\partial f}{\partial z}(z) = \begin{cases}
				  \begin{aligned}
				  	0 & \quad \text{ when } z \leqslant 0 \\
				  	1 & \quad \text{ when } z > 0 \\
				  \end{aligned}
               \end{cases}$ \\[12pt]
      \code{arctan}
	  & $f(z) = \arctan{(z)}$ & $\frac{\partial f}{\partial z}(z) = \frac{1}{1 + e^{z}}$ \\[10pt]
      \code{softplus}
	  & $f(z) =\ln{\left(1 + e^{z}\right)}$ & $\frac{\partial f}{\partial z}(z) = \frac{1}{1+e^{-z}}$ \\[10pt]
      \code{softmax}
	  & $f_i(\mathbf{z}) = \frac{e^{z_i}}{\sum\limits_k\left(e^{z_k}\right)} $ & $\frac{\partial f_i}{\partial z_j}(\mathbf{z}) = \begin{cases}
				  \begin{aligned}
				  	f_i(\mathbf{z}) \cdot (1-f_j(\mathbf{z})) & \quad \text{ when } i = j \\
				  	-f_j(\mathbf{z}) \cdot f_i(\mathbf{z}) & \quad \text{ when } i \neq j \\
				  \end{aligned}
               \end{cases}$ \\[12pt]
	 \hline
    \end{tabular}
  \end{center}
  \caption{Accepted activation functions and their derivatives in \fct{SensAnalysisMLP}, where $z$ refers to the input value of the neuron as described in Equation (\ref{eqn:output_l_output_l-1}) and $\mathbf{z}$ refers to the vector of input values of the neuron layer.}
  \label{tab:act_functs}
\end{table}

In order to show how \fct{SensAnalysisMLP} is used, we use a simulated dataset to train an MLP model of class \class{nn} (\pkg{RSNNS}). The dataset consists of a \class{data.frame} with 1500 rows of observations and four columns for three input variables (\code{X1}, \code{X2}, \code{X3}) and one output variable (\code{Y}). The input variables are random observations of a normal distribution with zero mean and standard deviation equal to 1. The output $Y$ is created following Equation (\ref{eqn:example_eqn}) based on $X_1$ and $X_2$:
\begin{equation}
\label{eqn:example_eqn}
	Y = (X_1)^2 - 0.5 \cdot X_2 + 0.1 \cdot \varepsilon
\end{equation} 
where $\varepsilon$ is random noise generated using a normal distribution with zero mean and standard deviation equal to 1. $X_3$ is given to the model for training and a proper fitted model would find no relation between $X_3$ and $Y$. 

\code{?NeuralSens::simdata} can be executed to obtain more information about the data. The library is loaded by executing the following code:
\begin{CodeChunk}
\begin{CodeInput}
R> library("NeuralSens")
\end{CodeInput}
\end{CodeChunk}
To test the functionality of the \fct{SensAnalysisMLP} function, \fct{mlp} function from \pkg{RSNNS} package trains a neural network model using the \code{simdata} dataset.
\begin{CodeChunk}
\begin{CodeInput}
R> library("RSNNS")
R> set.seed(150)
R> mod1 <- mlp(simdata[,c("X1","X2","X3")], simdata[,"Y"], maxit = 1000,
+    size = 10, linOut = TRUE)
\end{CodeInput}
\end{CodeChunk}
\fct{SensAnalysisMLP} is used to perform a sensitivity analysis to \code{mod1} using the same dataset as in training:
\begin{CodeChunk}
\begin{CodeInput}
R> sens <- SensAnalysisMLP(mod1, trData = simdata, output_name = "Y",
+    plot = FALSE)
\end{CodeInput}
\end{CodeChunk}
\code{sens} is a \class{SensMLP} object and methods of that class can be used to explore the sensitivity analysis:
\begin{CodeChunk}
\begin{CodeInput}
R> class(sens)
\end{CodeInput}
\begin{CodeOutput}
[1] "SensMLP"
\end{CodeOutput}
\begin{CodeInput}
R> summary(sens)
\end{CodeInput}
\begin{CodeOutput}
Sensitivity analysis of 3-10-1 MLP network.

Sensitivity measures of each output:
$Y
           mean        std meanSensSQ
X1 -0.005406908 1.94524276 1.94476390
X2 -0.485564931 0.06734504 0.49021056
X3 -0.003200699 0.02971083 0.02987535
\end{CodeOutput}
\end{CodeChunk}
\fct{summary} method prints the sensitivity measures of the output with regard to the inputs of the model. These measures are calculated using the sensitivities displayed when using the \fct{print} method described below. The \code{mean} column ($S_{ik}^{avg}$) shows the mean effect of the input variable on the output. The \code{std} column ($S_{ik}^{sd}$) shows the variance of the input variable's effect on the output along the input space. These columns provide information on the relation between inputs and output variables:
\begin{itemize}
	\item If both \code{mean} ($S_{ik}^{avg}$) and \code{std} ($S_{ik}^{sd}$) are near zero, it indicates that the output is not related to the input, because for all the training data the sensitivity of the output with regard to that input is approximately zero.
	\item If \code{mean} ($S_{ik}^{avg}$) is different from zero and \code{std} ($S_{ik}^{sd}$) is near zero, it indicates that the output has a linear relationship with the input, because for all the training data the sensitivity of the output with regard tothat input is approximately constant. 
	\item If \code{std} ($S_{ik}^{sd}$) is different from zero, regardless of the value of \code{mean} ($S_{ik}^{avg}$), it indicates that the output has a non-linear relationship with the input, because the relation between the output and the input vary depending on the value of the input.
\end{itemize} 

Setting an upperbound for \code{std} to be considered close to zero so that the relationship between output and input can be considered as linear is a non-trivial endeavor. The authors are working on a statistic to test whether the functional relationship between an input and an output variable can be considered linear and, if successful, it will be included in a future version of the package. 
\newpage
In the example, the \code{mean} and \code{std} values show:
\begin{itemize}
	\item $X_1$ has mean $\approx 0$ and standard deviation $\approx 2$. This means it has a non-constant, i.e., non-linear effect on the response variable.
	\item $X_2$ has mean $\approx 0.5$ and standard deviation $\approx 0$. This means it has a constant, i.e., linear effect on the response variable.
	\item $X_3$ has mean $\approx 0$ and standard deviation $\approx 0$. This means it has no effect on the response variable.
\end{itemize}

An input variable may be considered significant if their sensitivities $\evalat[\big]{s_{ik}}{\mathbf{x}_j}$ are significantly different from zero, whether they are positive or negative. In other words, a variable is considered to be significant when changes in the input variable produce significant changes in the output variable of the model. \cite{white_statistical_2001} conclude that the statistic $\left(S_{ik}^{sq}\right)^{2} = \frac{\sum_{j = 1}^{N} \left(\evalat[\big]{s_{ik}}{\mathbf{x}_j}\right)^2}{N}$ is a valid indicator to identify if a variable is irrelevant. Moreover, $S_{ik}^{sq}$ is a measure of the changes in the output due to local changes in the input. Thus, $S_{ik}^{sq}$ can be defined as a measure of the importance of the input variables from a perturbation analysis point of view, in the sense that small changes in that input will produce larger changes in the output.  

\class{SensMLP} class has also a \fct{print} method to show the sensitivities of the output with regard to the inputs evaluated in each of the rows of the \code{trData} component of the \code{sens} object. A second argument \code{n} may be used to specify how many rows to display (by default \code{n = 5}).
\begin{CodeChunk}
\begin{CodeInput}
R> print(sens, n = 2)
\end{CodeInput}
\begin{CodeOutput}
Sensitivity analysis of 3-10-1 MLP network.

  2000 samples

Sensitivities of each output (only 2 first samples):
$Y
              X1         X2           X3
[1,]  2.08642384 -0.4462707 -0.044063158
[2,] -0.34976334 -0.3547381  0.014188363
\end{CodeOutput}
\end{CodeChunk}
\subsection{Visualizing Neural Network Sensitivity Measures}\label{subsec:Vis_Sensitivity}

Sensitivity measures of the output variables are useful for quantitative analysis. However, it can be difficult to compare sensitivity metrics when a large number of input variables are used. In order to visualize information on the calculated sensitivities, the authors propose the following plots:

\begin{enumerate}
	\item Label plot representing the relationship between $S_{ik}^{avg}$ (x-axis) and $S_{k}^{std}$ (y-axis). 
	
	\item Bar plot that shows $S_{k}^{sq}$ for each input variable. 
	
	\item Density plot that shows the distribution of output sensitivities with regard to each input (\cite{muoz_variable_1998}):
	\begin{itemize}
		\item The narrow distribution of sensitivity values for \code{X2} (corresponding to a constant sensitivity) indicates a linear relationship between this input and the output of the neural net.
		\item The wide distribution of sensitivity values for \code{X1} (corresponding to a variable sensitivity) indicates a non-linear relationship between this input and the output.
	\end{itemize}
	When the height of at least one of the distributions is greater than 10 times the height of the smallest distribution, then an extra plot is created using the \fct{facet\_zoom} function of the \pkg{ggforce} package (\cite{ggforce}). These plots provides a better representation of the sensitivity distributions. 
\end{enumerate}

\begin{figure}[t!]
\centering
\includegraphics[width=0.75\linewidth, height=0.45\linewidth]{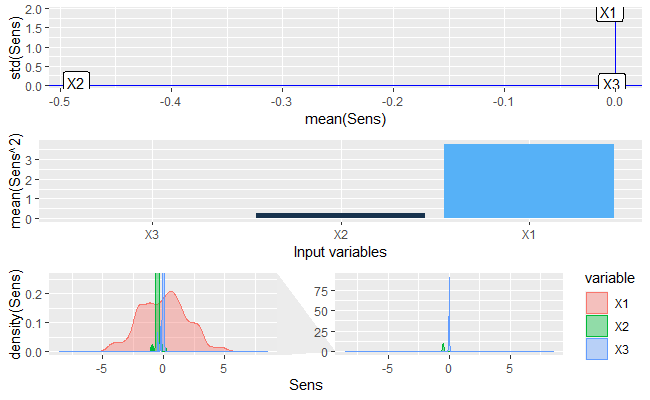}
\caption{\label{fig:exampleplots} Example from the \fct{SensitivityPlots} function showing plots specified in Section \ref{subsec:Vis_Sensitivity}. First plot shows the relation between the mean and standard deviation of the sensitivities, the second plot shows the square of the sensitivities and the third and fourth plots show the distribution of the sensitivities.}
\end{figure}

These plots can be obtained using the \fct{SensitivityPlots} function and a \class{SensMLP} object calculated using \fct{SensAnalysisMLP}. To obtain the plots of Figure \ref{fig:exampleplots}:
\begin{CodeChunk}
\begin{CodeInput}
R> SensitivityPlots(sens)
\end{CodeInput}
\end{CodeChunk}
Or they can be generated using the \fct{plot} method of the \class{SensMLP} object:
\begin{CodeChunk}
\begin{CodeInput}
R> plot(sens)
\end{CodeInput}
\end{CodeChunk}
In this case, the first plot of Figure \ref{fig:exampleplots} shows that \code{Y} has a negative linear relationship with \code{X2} ($std \approx 0$ and $mean < 0$), no relationship with \code{X3} ($std \approx 0$ and $mean \approx 0$) and a non-linear relationship with \code{X1} ($std$ different from $0$). The second plot shows that \code{X3} barely affects the response variable, being \code{X1} and \code{X2} the inputs with most effect on the output.

\subsection{Visualizing Neural Network Sensitivity over time}\label{subsec:Vis_Sensitivity_time}

A common application of neural networks is time series forecasting. Analyzing how sensitivities evolve over time can provide a better understanding of the effect of explanatory variables on the output variables. 

\fct{SensTimePlot} returns a sequence plot of the raw sensitivities calculated by the function \fct{SensAnalysisMLP}. The x-axis is related to a \code{numeric} or \code{Posixct}/\code{Posixlt} variable containing the time information of each sample. The y-axis is related to the sensitivities of the output with regard to each input. 

In order to show how this function can be used, the \code{DAILY_DEMAND_TR} dataset is used to create a model of class \class{train} from \pkg{caret} package (\cite{caret}). This dataset is similar to the \code{elecdaily} dataset from \pkg{fpp2} R package (\cite{fpp2}). However, \code{DAILY_DEMAND_TR} contains almost five years of daily data (\code{elecdaily} only one year), which makes it more suitable for training a neural network. It is composed of the following variables:
\begin{itemize}
	\item DATE: date of the sample, one per day from July, 2nd 2007 to November, 30th 2012.
	\item TEMP: mean daily temperature in ºC in Madrid, Spain.
	\item WD: working day, continuous parameter which represents the effect on the daily consumption of electricity as a percentage of the expected electricity demand of that day with regard to the demand of the reference day of the same week \cite{moral-carcedo_modelling_2005}. In this case, Wednesday is the reference day (WD$_{Wed} \approx 1$).
	\item DEM:  total daily electricity demand in GWh for Madrid, Spain.
\end{itemize} 

\begin{figure}[t!]
\centering
\includegraphics[width=0.7\linewidth, height=0.35\linewidth]{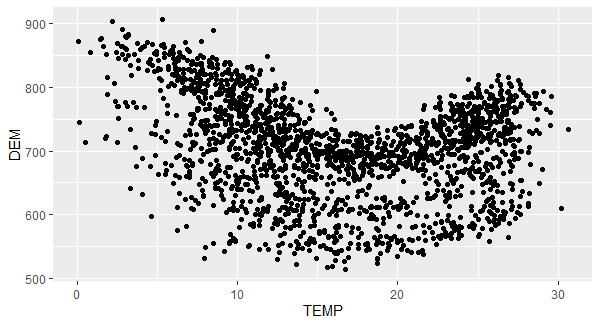}
\caption{Relation between the output variable, DEM, and the input TEMP of DAILY\_DEMAND\_TR database.} \label{fig:subfigDEMTEMP}
\end{figure}

The following code creates the plot in Figure \ref{fig:subfigDEMTEMP}:  
\begin{CodeChunk}
\begin{CodeInput}
R> library("ggplot2")
R> ggplot(DAILY_DEMAND_TR) + geom_point(aes(x = TEMP, y = DEM))
\end{CodeInput}
\end{CodeChunk}
Figure \ref{fig:subfigDEMTEMP} shows the relationship between the electricity demand and the temperature. A non-linear effect can be observed, where the demand increases for low temperatures (due to heating systems) and for high temperatures (due to air conditioners). 

\begin{figure}[t!]
\centering
\includegraphics[width=\linewidth, height=0.55\textwidth]{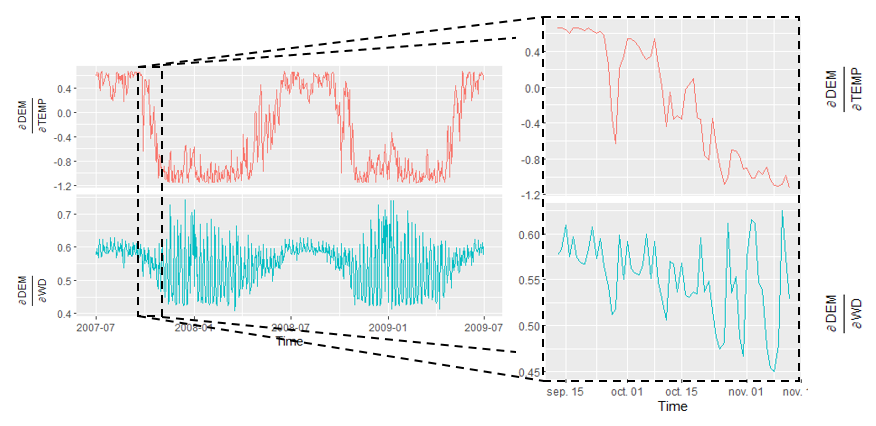}
\caption{\label{fig:sensovertime} Example from the \fct{SensTimePlot} function, showing how the sensitivities for each of the inputs evolve over time.}
\end{figure}

The following code scales the data, create a \code{train} neural network model and apply the \fct{SensTimePlot} function to two years of the data:
\begin{CodeChunk}
\begin{CodeInput}
R> DAILY_DEMAND_TR[,4] <- DAILY_DEMAND_TR[,4]/10
R> DAILY_DEMAND_TR[,2] <- DAILY_DEMAND_TR[,2]/100
R> library("caret")
R> set.seed(150)
R> mod2 <- train(form = DEM~TEMP + WD, data = DAILY_DEMAND_TR, 
+    method = "nnet", linout = TRUE, maxit = 250, metric = "RMSE", 
+    tuneGrid = data.frame(size = 5, decay = 0.1),
+    preProcess = c("center","scale"), trControl = trainControl())
R> SensTimePlot(mod2, DAILY_DEMAND_TR[1:(365*2),], output_name = "DEM",
+    date.var = DAILY_DEMAND_TR[1:(365*2),1], facet = TRUE)
\end{CodeInput}
\end{CodeChunk}
Figure \ref{fig:sensovertime} shows that the temperature variable has a seasonal effect on the response variable. In summer, the temperature is higher and cooling systems demand more electricity, therefore the demand is directly proportional to the temperature. In winter, the temperature is lower and heating systems demand more electricity, hence the demand is inversely proportional to the temperature. The sensitivity of the output with regard to \code{WD} has also a seasonal effect, with higher variance in winter than in summer and greater sensitivity in weekends. Figure \ref{fig:sensovertime} can also be generated using the \fct{plot} method of a \class{SensMLP} object:
\begin{CodeChunk}
\begin{CodeInput}
R> sens2 <- SensAnalysisMLP(mod2, trData = DAILY_DEMAND_TR[1:(365*2),], 
+    output_name = "DEM", plot = FALSE)
R> plot(sens2, plotType = "time", facet = TRUE, 
+    date.var = DAILY_DEMAND_TR[1:(365*2),1])
\end{CodeInput}
\end{CodeChunk}

\subsection{Visualizing the Neural Network Sensitivity relation as a function of the input values}\label{subsec:Vis_Sensitivity_inputs}

Sometimes it is useful to know how the value of the input variables affects the sensitivity of the response variables. The \fct{SensFeaturePlot} function produces a violin plot to show the probability density of the output sensitivities with regard to each input. It also plots a  jitter strip chart for each input, where the width of the jitter is controlled by the density distribution of the data (\cite{ggforce}). The color of the points is proportional to the value of each input variable, which display whether the relation of the output with the input is relatively constant within a range of input values.

\begin{figure}[t!]
\centering
\includegraphics[width=0.7\linewidth, height=0.5\linewidth]{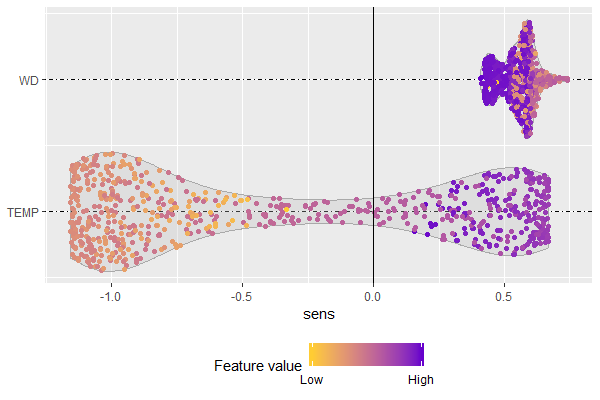}
\caption{\label{fig:sensfeatplot} Example from the \fct{SensFeaturePlot} function, showing the relationship of the sensitivities with the value of the inputs.}
\end{figure}

The following code produce the plot of Figure \ref{fig:sensfeatplot}:
\begin{CodeChunk}
\begin{CodeInput}
R> SensFeaturePlot(mod2, fdata = DAILY_DEMAND_TR[1:(365*2),])
\end{CodeInput}
\end{CodeChunk}

It can also be generated using the \fct{plot} method of a \class{SensMLP} object:
\begin{CodeChunk}
\begin{CodeInput}
R> plot(sens2, plotType = "features")
\end{CodeInput}
\end{CodeChunk}
In accordance with the information extracted from Figure \ref{fig:sensovertime}, Figure \ref{fig:sensfeatplot} shows that the sensitivity of the output with regard to the temperature is negative when the temperature is low and positive when the temperature is high. It also shows that the sensitivity of the output with regard to \code{WD} is higher in the weekends (lower values of \code{WD}).

\subsection{Extending package functionalities to other MLP models}\label{subsec:package_extend}

The current version of \pkg{NeuralSens} package (version 1.0.0), includes methods of \fct{SensAnalysisMLP} function  for \class{nn} (\pkg{neuralnet}), \class{nnet} (\pkg{nnet}), \class{H2ORegressionModel} and \class{H2OMultinomialModel} (\pkg{h2o}), \class{mlp} (\pkg{RSNNS}), \class{list} (\pkg{neural}), \class{nnetar} (\pkg{forecast}) and \class{train} (\pkg{caret}) (only if the object inherits the class attribute from another of the available packages). Additionally, a \class{numeric} method is available to perform sensitivity analysis of a new neural network model using only the weights of the model, its neural structure, and the activation function of the layers and their derivatives. 
The first information that must be extracted from the model are the weights of the connections between layers. These weights must be passed to the first argument of the \fct{SensAnalysisMLP} function as a \class{numeric} \class{vector}, concatenating the weights of the layers in order from the first hidden layer ($l = 2$) to the output layer ($l = L$). The bias weight should be added to the vector before the weights of the same layer, following the equation below:
\begin{equation} 
\label{eqn:neuralweightexample}
	wts = [b^2,w^2_{11},w^2_{21},...,w^2_{n^2n^1},b^3,w^3_{11},...,b^L,w^L_{11},...,w^L_{n^Ln^{L-1}}]
\end{equation}
If the model has no bias, the bias weights must be set to 0 ($b^l = 0$).

The second information is the neural structure of the model. The structure of the model must be passed to the \code{mlpstr} argument as a \class{numeric} \class{vector} equal in length to the number of layers in the network. Each number specifies the number of neurons in each layer, starting with the input layer and ending with the output layer:

The last information that must be provided are the activation functions of each layer and their derivatives. If the activation function of a layer is one of those provided by the package (shown in Table \ref{tab:act_functs}), the function can be specified using its name. If the activation function is not one of those provided in the package, it should be passed as a function. The same applies to the derivative of the activation function. The activation function  $\Phi^l(\mathbf{z}^l)$ of a layer $l$ and its derivative $\frac{\partial(\Phi^l(\mathbf{z}^l))}{\partial(\mathbf{z}^l)}$ must meet the following conditions:
\begin{itemize}
	\item $\Phi^l(\mathbf{z}^l)$ must return a \class{vector} with the same length as $\mathbf{z}^l$. The activation function of each neuron may be different, as long as this condition is met:
	\begin{align}
		\Phi^l(\mathbf{z}^l) &= \begin{bmatrix}
           \phi^l_1(z^l_{1}) \\
           \phi^l_2(z^l_{2}) \\
           \vdots \\
           \phi^l_{n^l}(z^l_{n^l})
         \end{bmatrix}
	\end{align}
	\item $\frac{\partial(\Phi^l(\mathbf{z}^l))}{\partial(\mathbf{z}^l)}$ must return a square \class{matrix} with the derivative of $\Phi^l(\mathbf{z}^l)$ with regard to each component of $\mathbf{z}^l$:
	\begin{equation}
		\frac{\partial(\Phi^l(\mathbf{z}^l))}{\partial(\mathbf{z}^l)} = \left[ {\begin{array}{cccc}
	\frac{\partial \phi^l_1}{\partial z^{l}_1}\left(z^l_1\right) & \frac{\partial \phi^l_2}{\partial z^{l}_1}\left(z^l_1\right) & \cdots & \frac{\partial \phi^l_{n^l}}{\partial z^{l}_1}\left(z^l_1\right) \\
	\frac{\partial \phi^l_1}{\partial z^{l}_2}\left(z^l_2\right) & \frac{\partial \phi^l_2}{\partial z^{l}_2}\left(z^l_2\right) & \cdots & \frac{\partial \phi^l_{n^l}}{\partial z^{l}_2}\left(z^l_2\right) \\
	\vdots & \vdots & \ddots & \vdots \\
	\frac{\partial \phi^l_1}{\partial z^{l}_{n^{l}}}\left(z^l_{n^{l}}\right) & \frac{\partial \phi^l_2}{\partial z^{l}_{n^{l}}}\left(z^l_{n^{l}}\right) & \cdots & \frac{\partial \phi^l_{n^l}}{\partial z^{l}_{n^{l}}}\left(z^l_{n^{l}}\right) \\
	\end{array} } \right]
	\end{equation}
\end{itemize}
Examples of how to use the \fct{SensAnalysisMLP} function with new packages can be found in Appendix \ref{app:ext_pack_func_impl}.

\subsection{Effect of network structure and training conditions} \label{subsec:robustness}

An important advantage of sensitivity analysis based on partial derivatives is the robustness of the analysis results regardless of the model's neural structure. Other methods such as Olden rely heavily on the neural structure and the initial starting weights. A similar analysis to the one performed on \fct{olden} in \cite{beck_neuralnettools_2018} has been performed on the \fct{SensAnalysisMLP} function. To observe the effect of the neural structure on the sensitivity metrics, these metrics have been calculated for models with 1, 10 and 20 neurons in the hidden layer. For each neuron level, 50 models with different random initial weights are trained. If the neural structure and the initial starting weights have no effect on the sensitivity metrics, these metrics should be the same for all the models. \code{simdata} dataset is used to train the models. 

\begin{figure}[t!]
\centering
\includegraphics[width=0.8\textwidth, height=0.7\textwidth]{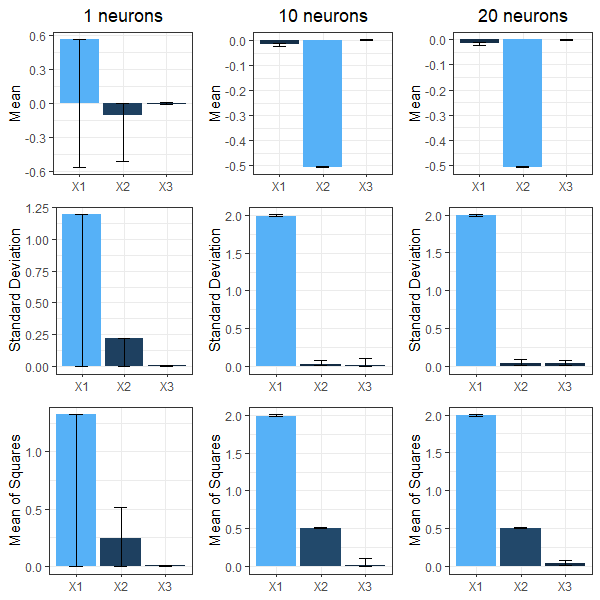}
\caption{\label{fig:robust_analysis} Robustness analysis of sensitivity metrics for three neural network models with different number of neurons in the hidden layer. Fifty different models for each neuron level are trained with random initial weights. \code{geom\_bar} shows the mean value of the sensitivity metrics and \code{geom\_errorbar} shows the minimum and maximum value of the sensitivity metrics.}
\end{figure}

Figure \ref{fig:robust_analysis} shows the mean value of the sensitivity metrics from the 50 models for each neural structure. It also shows the minimum and maximum value of the metric to display the effect of the neural structure and the initial weights values. An important conclusion that can be derived from Figure \ref{fig:robust_analysis} is that with enough neurons in the hidden layer, i.e., if the model can predict the output with enough precision; variance of sensitivity metrics is negligible compared to the value of the metric. 

\section{Further examples and comparison with other methods} \label{sec:further examples}

This section contains several examples in which the functions of the \pkg{NeuralSens} package are compared with similar functions from other \proglang{R} packages. Section \ref{subsec:mlp_class}
trains an MLP for classification to compare \fct{SensAnalysisMLP} with \fct{olden}, \fct{garson} \citep{beck_neuralnettools_2018} and \fct{plot\_explanations} \citep{lin_pedersen_understanding_2019}. Section \ref{subsec:mlp_reg} trains an MLP for regression to compare \fct{SensAnalysisMLP} and \fct{SensFeaturePlot} with \fct{lekprofile} \citep{beck_neuralnettools_2018} and \fct{partial} \citep{pdp}.

Topics such as data pre-processing or network architecture should be considered before model development. Discussions about these ideas have been already held (\cite{cannas_data_2006}, \cite{amasyali_review_2017}, \cite{maier_neural_1999}, \cite{lek_application_1996}) and are beyond the scope of this paper.

\subsection{Multilayer perceptron for classification} \label{subsec:mlp_class}

\begin{figure}[t!]
\centering
\begin{subfigure}[h]{0.48\textwidth}
\includegraphics[width=0.8\linewidth, height=0.8\textwidth]{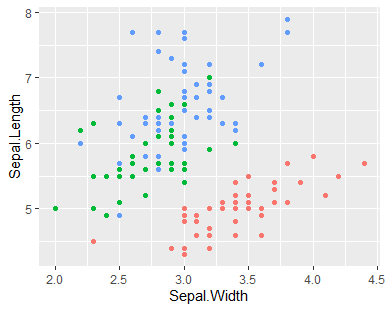}
\caption{Sepal-related variables} \label{subfig:sepal_ggplot}
\end{subfigure}
\hspace*{\fill} 
\begin{subfigure}[h]{0.48\textwidth}
\includegraphics[width=\linewidth, height=0.8\textwidth]{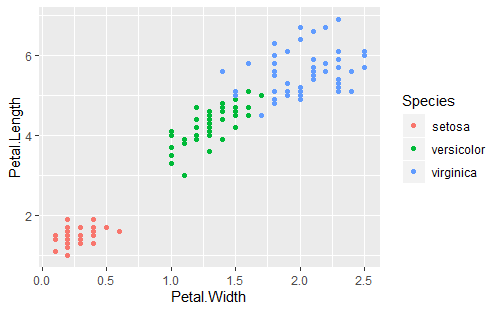}
\caption{Petal-related variables} \label{subfig:petal_ggplot}
\end{subfigure}
\caption{\label{fig:SepalPetal} (\ref{subfig:sepal_ggplot}) \code{geom\_point} plot representing the variables \code{Sepal.Length} and \code{Sepal.Width}, (\ref{subfig:petal_ggplot}) \code{geom\_point} plot representing the variables \code{Petal.Length} and \code{Petal.Width}}
\end{figure}

In this example a multilayer perceptron is trained using the well-known \code{iris} dataset included in R. Figure \ref{fig:SepalPetal} shows two scatterplots comparing the petal-related and sepal-related variables of the flowers in the dataset. It can be seen that \code{setosa} species have a smaller petal size than the other two species and a shorter sepal. It also shows that \code{virginica} and \code{versicolor} species have a similar sepal size, but the latter has a slightly smaller petal size.

The \fct{train} function from the \pkg{caret} package creates a new neural network model to predict the species of each flowers based on petal and sepal dimensions. 
\begin{CodeChunk}
\begin{CodeInput}
R> set.seed(150)
R> mod3 <- caret::train(Species~., data = iris, preProcess = c("center",
+    "scale"), method = "nnet", linout = TRUE, trControl = trainControl(), 
+    tuneGrid = data.frame(size = 5, decay = 0.1), metric = "Accuracy")
\end{CodeInput}
\end{CodeChunk}
\fct{SensAnalysisMLP} function calculates the sensitivities of the model, providing information of the relationships between each output class and each input variable.
\begin{CodeChunk}
\begin{CodeInput}
R> sens4 <- SensAnalysisMLP(mod3)
R> summary(sens4)
\end{CodeInput}
\begin{CodeOutput}
Sensitivity analysis of 4-5-3 MLP network.

Sensitivity measures of each output:
$setosa
                    mean        std meanSensSQ
Sepal.Length -0.01346027 0.02698245 0.03007287
Sepal.Width   0.03116669 0.05055589 0.05924712
Petal.Length -0.07133925 0.10172196 0.12396638
Petal.Width  -0.06846395 0.09654030 0.11808983

$versicolor
                     mean        std meanSensSQ
Sepal.Length  0.035833076 0.02252307 0.04228375
Sepal.Width  -0.002101449 0.03925762 0.03918294
Petal.Length -0.099221391 0.17568963 0.20126091
Petal.Width  -0.093949654 0.16300239 0.18766775

$virginica
                    mean        std meanSensSQ
Sepal.Length -0.01803139 0.02742334 0.03274380
Sepal.Width  -0.04199856 0.05663125 0.07035338
Petal.Length  0.20543515 0.28413548 0.34985476
Petal.Width   0.19731097 0.27236735 0.33559057

\end{CodeOutput}
\end{CodeChunk}

\begin{figure}[t!]
\centering
	\begin{tabular}{cc}
	\begin{subfigure}[h]{0.5\textwidth}
	\centering
	\includegraphics[width=0.9\linewidth, height=1.2\textwidth]{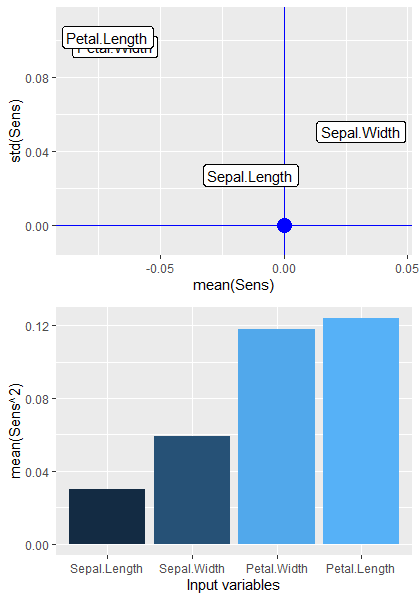}
    \caption{Model 3, \fct{SensitivityPlots}} \label{subfig:mod3_sensplot}
	\end{subfigure}
	&
		\begin{tabular}{c}
            \begin{subfigure}[h]{0.45\textwidth}
                \centering
                \includegraphics[width=\linewidth, height=0.65\textwidth]{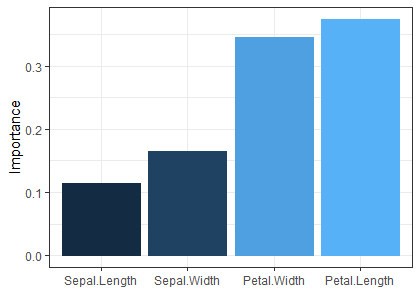}
                \caption{Model 3, \fct{garson}} \label{subfig:mod3_garson}
            \end{subfigure}\\
            \begin{subfigure}[h]{0.45\textwidth}
                \centering
                \includegraphics[width=\linewidth, height=0.65\textwidth]{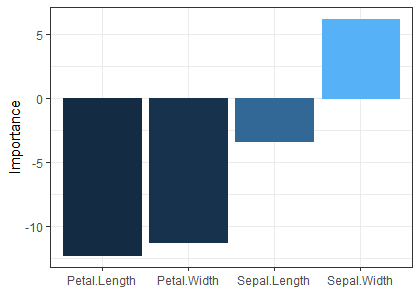}
                \caption{Model 3, \fct{olden}} \label{subfig:mod3_olden}
            \end{subfigure}
        \end{tabular}\\
	\end{tabular}
	\caption{\label{fig:mod3_sensAnalysis}(\ref{subfig:mod3_sensplot}) Global sensitivity measures plots of \code{mod3} for the \code{iris} dataset using \fct{SensitivityPlots}, (\ref{subfig:mod3_garson}) variable importance using \fct{garson} from \pkg{NeuralNetTools}, (\ref{subfig:mod3_olden}) variable importance using \fct{olden} from \pkg{NeuralNetTools}.}
\end{figure}

The sensitivity metrics for each of the output provides information on how the neural network uses the data to predict the output:
\begin{itemize}
	\item The \code{setosa} class has a greater probability when \code{Petal.Length}, \code{Petal.Width} and \code{Sepal.Length} variables decrease, or the \code{Sepal.Width} variable increases.
	\item The \code{versicolor} class has a greater probability when \code{Petal.Length} and \code{Petal.Width} variables decrease, or the \code{Sepal.Length} variable increases.
	\item The \code{virginica} class has a greater probability when the \code{Petal.Length} and \code{Petal.Width} variables increase, and \code{Sepal.Length} and \code{Sepal.Width} variables decrease.
\end{itemize}

This information corresponds to what is observed in Figure \ref{fig:SepalPetal}, where \code{setosa} class is characterized by a low value of \code{Petal.Length}, \code{Petal.Width} and \code{Sepal.Length} variables, and \code{versicolor} and \code{virginica} classes are differentiated by the value of the \code{Petal.Length} and \code{Petal.Width} variables. 

\fct{garson} and \fct{olden} method from the \pkg{NeuralNetTools} package (\cite{beck_neuralnettools_2018}) provide information on input importance. As they provide information related to the first output neuron, the comparison with \fct{SensAnalysisMLP} is done using the sensitivity measures for the first output class.
\begin{CodeChunk}
\begin{CodeInput}
R> SensitivityPlots(sens, der = FALSE, output = "setosa")
R> garson(mod3)
R> olden(mod3)
\end{CodeInput}
\end{CodeChunk}
Figure \ref{subfig:mod3_sensplot} shows the sensitivity metrics calculated by \fct{SensAnalysisMLP}, Figure \ref{subfig:mod3_garson} shows \fct{garson}'s importance metrics for the input variables and Figure \ref{subfig:mod3_olden} shows \fct{olden}'s importance metrics for the input variables. The mean value of the sensitivities in the top chart of Figure \ref{subfig:mod3_sensplot} is similar to \fct{olden}'s metrics observed in \ref{subfig:mod3_olden}, and the $S_{ik}^{sq}$ values in the barplot of Figure \ref{subfig:mod3_sensplot} are similar to \fct{garson}'s values observed in Figure \ref{subfig:mod3_garson}. It must be noted that values from \fct{SensAnalysisMLP} are more robust against changes in the neural structure and initial weights as stated in Section \ref{subsec:robustness} and \cite{beck_neuralnettools_2018}.

\begin{figure}[t!]
\centering
\includegraphics[width=0.8\linewidth, height = 0.6\textwidth]{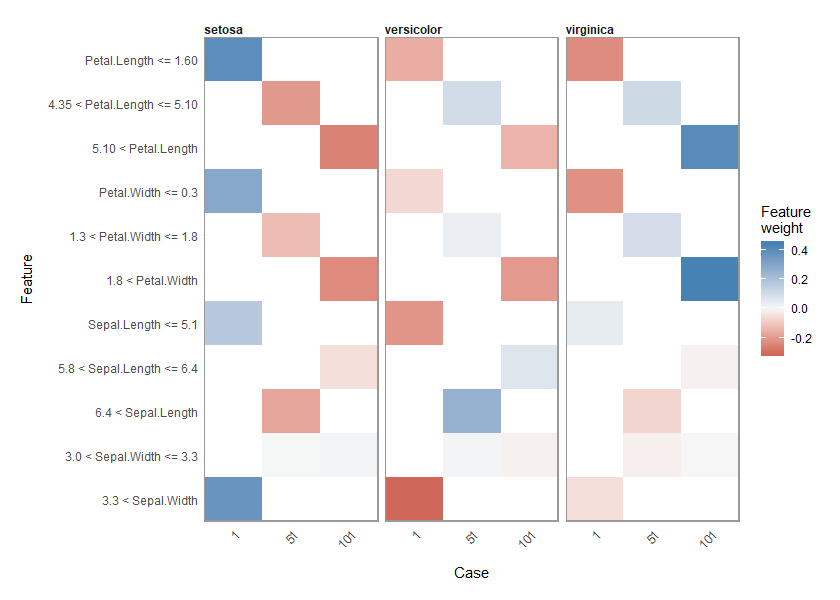}
\caption{\label{fig:lime} Facetted heatmap-style plots generated by applying the \fct{plot\_explanations} function to \code{mod3} for three selected samples of \code{iris} dataset. Each plot represents the contribution (positive or negative) of each feature to the probability of a specific class of \code{iris[,"Species"]} for the specified data samples..}
\end{figure}

The \pkg{lime} (\cite{lin_pedersen_understanding_2019}) package can also be used to obtain information on the neural network model. In this case, \fct{lime} and \fct{explain} functions train a decision tree model using the entire dataset to interpret how the neural network predicts the class of three different samples (one for each iris species) using all the features in the training dataset. \fct{plot\_explanations} function shows graphically the information given by the decision tree.
\begin{CodeChunk}
\begin{CodeInput}
R> library("lime")
R> plot_explanations(explain(iris[c(1,51,101),], lime(iris, mod3),
+    n_labels = 3, n_features = 4))
\end{CodeInput}
\end{CodeChunk}

Figure \ref{fig:lime} confirms the relationships between the inputs and the output variable. However, this method does not provide a quantitative measure for the importance of each input. Due to the lack of quantitative measures for  input importance this method can not be directly compared to the other methods exposed in this section (Figure \ref{fig:mod3_sensAnalysis}). 

Sometimes it may be more interesting to obtain global importance measures instead of measures for each output variable. \pkg{NeuralSens} allows us to obtain global measures using the \fct{CombineSens} function. It computes the sensitivity measures of the whole model following Equations (\ref{eqn:global_M_AV_S}), (\ref{eqn:global_STD_S}) and (\ref{eqn:global_M_SQ_AV_S}). These global measures are an indicator of how much, on average, the output probabilities change when an input variable changes.

\begin{CodeChunk}
\begin{CodeInput}
R> summary(CombineSens(sens))
\end{CodeInput}
\begin{CodeOutput}
Sensitivity analysis of 4-5-3 MLP network.

Sensitivity measures of each output:
$Combined
                     mean        std meanSensSQ
Sepal.Length  0.001447138 0.03545616 0.03503348
Sepal.Width  -0.004311108 0.05770054 0.05626114
Petal.Length  0.011624836 0.24404886 0.22502735
Petal.Width   0.011632454 0.23246026 0.21378272
\end{CodeOutput}
\end{CodeChunk}

\subsection{Multilayer perceptron for regression} \label{subsec:mlp_reg}

The \code{Boston} dataset from the \pkg{MASS} (\cite{ripley_mass_2019}) package is used to train an \code{nnet} (\cite{nnet}) model. This dataset contains information collected by the U.S. Census Service on housing in the suburbs of Boston (run \code{?MASS::Boston} to obtain more information about the dataset). 

The objective of the model is to predict the nitric oxides concentration (parts per 10 million), stored in the \code{nox} variable. The input variables of the model are \code{zn} (proportion of residential land zoned for lots over 25,000 sq.ft.), \code{rad} (index of accessibility to radial highways) and \code{lstat} (lower status of the population (percent)). \fct{scale} function standardizes the input variables. \fct{SensFeaturePlot}, \fct{SensAnalysisMLP}, \fct{lekprofile} (\pkg{NeuralNetTools} \cite{beck_neuralnettools_2018}) and \fct{pdp} (\pkg{pdp} \cite{pdp}) functions analyze the relationships of the output with regard to the inputs.
\begin{CodeChunk}
\begin{CodeInput}
R> data("Boston", package = "MASS")
R> Boston <- as.data.frame(scale(Boston[,c("zn","rad","lstat","nox")]))
R> set.seed(150)
R> mod4 <- nnet::nnet(nox ~ ., data = Boston, size = 15, decay = 0.1, 
+    maxit = 150)
R> lekprofile(mod4, group_vals = 6)
R> lekprofile(mod4, group_vals = 6, group_show = TRUE)
R> library("pdp")
R> pdps <- list()
R> for (i in 1:3) {
+    pdps[[i]] <- autoplot(partial(mod4, pred.var = names(Boston)[i],
+      train = Boston, ice = TRUE), train = Boston, 
+      center = TRUE, alpha = 0.2, rug = TRUE) +
+   theme_bw() + ylab("nox")
+  }
R> gridExtra::grid.arrange(grobs = pdps, nrow = 1)
R> SensFeaturePlot(mod4, fdata = Boston, output_name = "nox")
R> SensAnalysisMLP(mod4, trData = Boston, output_name = "nox")
\end{CodeInput}
\end{CodeChunk}
\begin{figure}[t!]
\centering
\begin{subfigure}[h]{0.8\textwidth}
\includegraphics[width=\linewidth, height=0.32\textwidth]{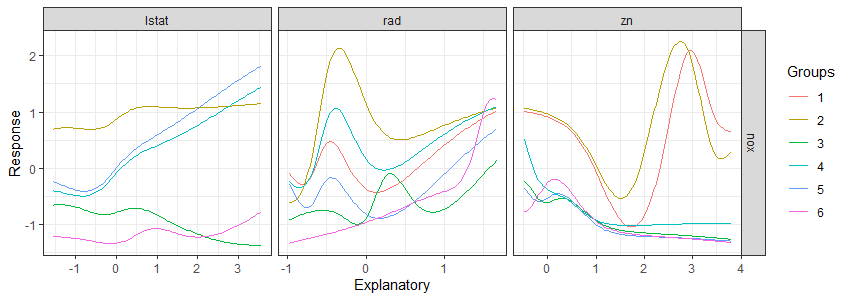}
    \caption{Model 4, \fct{lekprofile}} \label{subfig:mod4_lekprofile}
\end{subfigure}
\begin{subfigure}[h]{0.8\textwidth}
\includegraphics[width=\linewidth, height=0.32\textwidth]{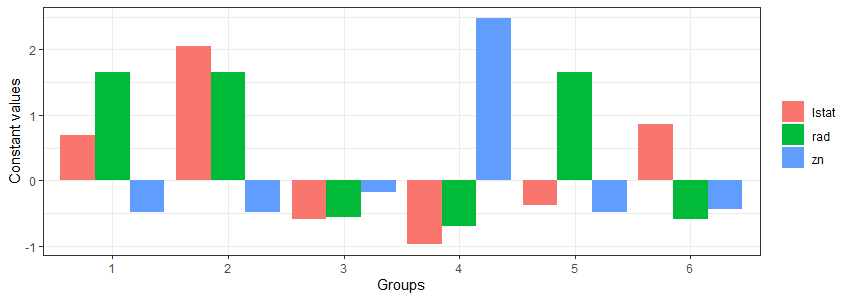}
    \caption{Cluster grouping of \fct{lekprofile}} \label{subfig:mod4_groupshow}
\end{subfigure}
\begin{subfigure}[h]{0.8\textwidth}
\includegraphics[width=\linewidth, height=0.32\textwidth]{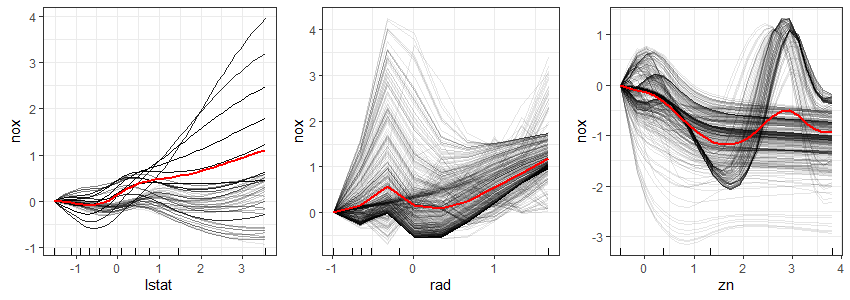}
\caption{Model 4, \code{pdp} and \code{ICE} plots}
\label{subfig:pdp}
\end{subfigure}
\begin{subfigure}[h]{0.48\textwidth}
\includegraphics[width=\linewidth, height=0.53\textwidth]{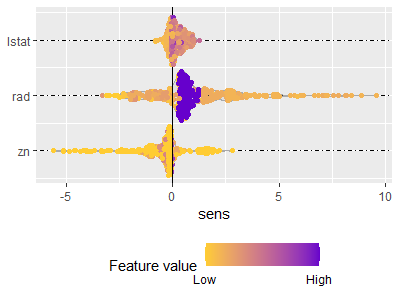}
\caption{Model 4, \fct{SensFeaturePlot}}
 \label{subfig:feat}
\end{subfigure}
\begin{subfigure}[h]{0.48\textwidth}
\includegraphics[width=\linewidth, height=0.53\textwidth]{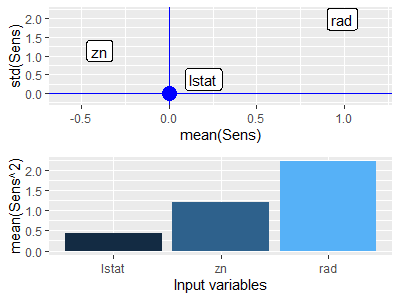}
\caption{Model 4, \fct{SensAnalysisMLP} plots}
 \label{subfig:sensans4}
\end{subfigure}
\caption{\label{fig:lekpdpfeatcomparison} (\ref{subfig:mod4_lekprofile}) Sensitivity analysis of a neural network using \fct{lekprofile} from \pkg{NeuralNetTools}. (\ref{subfig:mod4_groupshow}) Values at which explanatory variables are held constant for each cluster in \fct{lekprofile}. \ref{subfig:pdp} Partial dependence plots (red) and Individual Conditional Expectation plots (black) of \code{mod4}. \ref{subfig:feat} \fct{SensFeaturePlot} applied to \code{mod4}. \ref{subfig:sensans4} \fct{SensAnalysisMLP} applied to \code{mod4}.}
\end{figure}

Figures \ref{subfig:mod4_lekprofile} and \ref{subfig:mod4_groupshow} display the results of Lek's profile method from the \pkg{NeuralNetTools} package. To prevent the analysis of non-representative scenarios in the input dataset, a k-means clustering with 6 clusters has been applied to the dataset. Each subplot in Figure \ref{subfig:mod4_lekprofile} shows the evolution of the output variable when varying the input variable of interest across a range of values corresponding to the center of the k-means clusters. The other inputs remain constant in their values in the center of each k-means cluster. Figure \ref{subfig:mod4_groupshow} shows the value of the input variables at each of the cluster centers.

Figure \ref{subfig:pdp} shows the partial dependence plots (pdp) and Individual Conditional Expectaction (ICE) plots of \code{mod4} output with regard to each input variable. An ICE curve is calculated by maintaining the input variable of interest $x_i$ at a value $x_{ik}$, where $x_{ik}$ is the value of $x_i$ in the $k$ row of the dataset, and varying all other inputs across their values in the dataset. For a given dataset with $N$ samples, there would be $N$ ICE curves for each input variable. The PDP curve can be calculated as the mean of these ICE curves. PDP shows the marginal effect a given input variable has on the response variable of the neural network model, averaging the effects of the rest of the input variables. 

Calculating all ICE curves shows how the output variable change in the entire input space of the dataset. This comes at a large computational cost, since the number of curves that must be calculated are directly proportional to the number of samples and number of input variables. The computational time can be reduced by calculating only the PDP or a reduced number of ICE curves, but in that case some important scenarios might be ignored. 

\fct{SensFeaturePlot} function performs an analysis similar to Lek's and \code{pdp} by plotting the sensitivity of the output with regard to the input colored proportionally to the value of the input variable. On the one hand, \fct{lekprofile} function indicates that \code{lstat} and \code{rad} have a direct relationship with the output and \code{zn} has an inverse relationship with the output. Figure \ref{subfig:mod4_lekprofile} also suggests that all the input variables have a non-linear relationship with the output. On the other hand, Figure \ref{subfig:feat} shows that the sensitivity of the output has an approximately linear relationship with \code{zn}, and a non-linear relationship with the other two inputs. In this case, the \fct{lekprofile} function gives more information about the model, but it can be difficult to understand how the value of the input variables in each group affects the output variable.

\fct{SensAnalysisMLP} can be used to obtain more information on variable importance and relationships between variables. In this case, the \code{rad} variable affects the output the most, information which can be difficult to extract from the other functions.

\subsection{Computational cost}

As the size of neural network models increase exponentially to solve more complex tasks, performing a sensitivity analysis of a model could become an intensive computational task. Sensitivity analysis using partial derivatives requires matrix calcules where the size of the matrixes is directly proportional to the size of the hidden layers. As the number of neurons in hidden layers increase, the time to perform these calculations grows rapidly.

A comparison of how much time is required by a sensitivity analysis using the different methods included in Section \ref{sec:further examples} has been performed. This comparison has been carried out using the \code{YearPredictionMSD} dataset. This dataset consists of 90 input variables and 515345 samples. The authors propose to measure the computational time when varying the number of input variables, the number of samples and the size of the hidden layer of a single hidden layer MLP model. 

\begin{figure}[t!]
\centering
\begin{subfigure}[h]{\textwidth}
\includegraphics[width=\linewidth, height=0.37\textwidth]{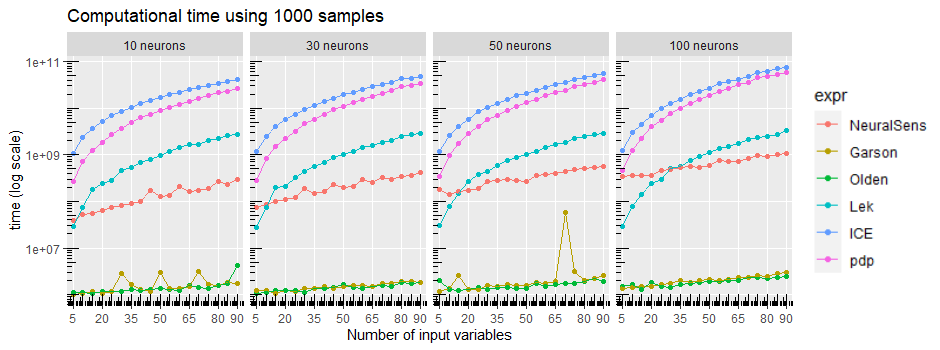}
\caption{Computational time using 1000 training samples}
\end{subfigure}
\begin{subfigure}[h]{\textwidth}
\includegraphics[width=\linewidth, height=0.37\textwidth]{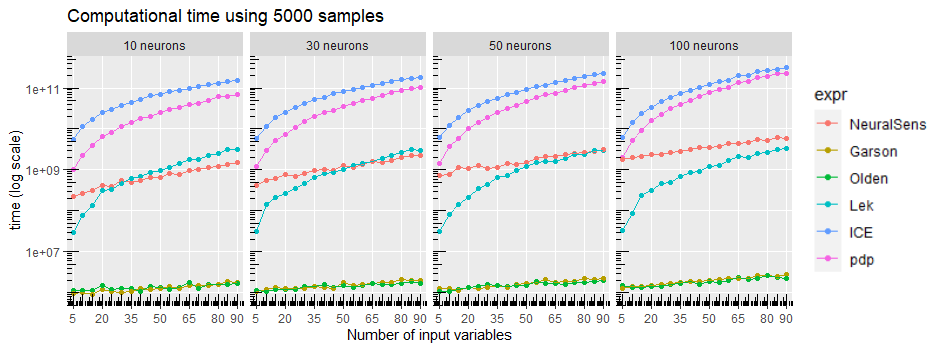}
\caption{Computational time using 5000 training samples}
\end{subfigure}
\begin{subfigure}[h]{\textwidth}
\includegraphics[width=\linewidth, height=0.37\textwidth]{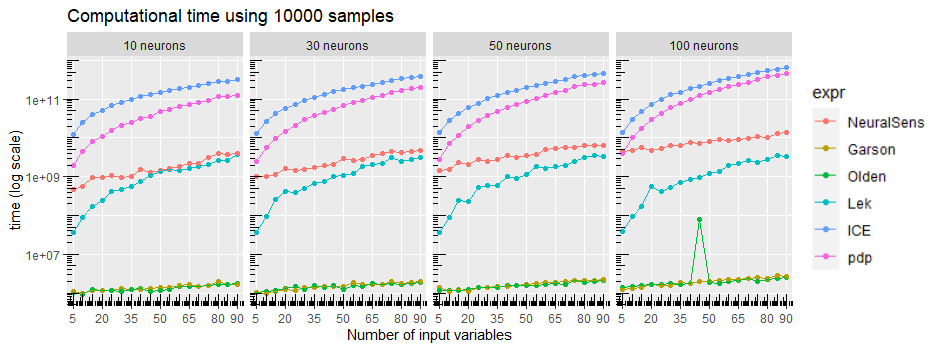}
\caption{Computational time using 10000 training samples}
\end{subfigure}
\caption{\label{fig:compt_time} Computational time of the different sensitivity analysis methods with different number of training samples (1000, 5000 and 10000 training samples), number of input variables (from 5 to 90 input variables) and number of neurons in the hidden layer (10, 30, 50 and 100 neurons).}
\end{figure}

This analysis has been performed on a computer with the following specs: processor Intel(R) Core(TM) i7-8700 @3.20 GHz, 32 GB of RAM memory, R version 3.6.3 (2020-02-29), platform x86\_64-w64-mingw32/x64 (64-bit) and running under Windows 10 x64 (build 18362).

Figure \ref{fig:compt_time} shows the computational time of each function varying the number of training samples, the number of input variables and the number of neurons in the hidden layer. Some conclusions can be reached from this figure:
\begin{itemize}
\item The fastest functions are the \fct{olden} and \fct{garson} functions from the \pkg{NeuralNetTools} package, because they only perform a sum of the weight matrices of the model. As it does not depend of the number of samples, the computational time is directly proportional to the size of the neural network layers (input and hidden layers).
\item \fct{lekprofile} from \pkg{NeuralNetTools} and \fct{SensAnalysisMLP} from \pkg{NeuralSens} need of the similar computational times.h \fct{SensAnalysisMLP} is affected byFthe number of neurons in the hidden layer and the number of samples, as the size of the matrixes increases with the number of neurons and the number of matrixes multiplication increase with the number of samples. \fct{lekprofile} is more affected by the number of input variables, because the number of curves to be created increase with the number of input variables. However, due to the fact that it uses only a fixed number of scenarios it does not depend on the number of samples. Since the number of neurons in the hidden layer barely affects the computational time to predict the output variable in each scenario, this parameter does not affect the computational time of lekprofile.
\item The slowest function is the \fct{partial} function from the \pkg{pdp} package when calculating all ICE curves. Calculating all ICE curves instead of only the pdp curves adds a noticeable amount of computational time. However, if the form of the ICE curves is not constant throughout the samples of the dataset (as in Figure \ref{subfig:pdp}), showing only the pdp curve gives misleading information as the form of the pdp curve does not resemble all the ICE curves. The computational time for \fct{partial} is directly proportional to the number of samples and the number of input variables, because the number of curves to be calculated increase exponentially as these parameters increase.
\end{itemize}

In addition to these conclusions, it must be mentioned that this analysis has been performed using a model with only one output variable. If there were several output variables, to obtain analogous information as \fct{SensAnalysisMLP} the other functions must be called once for each output. Because of this, the computational time of all the functions except \fct{SensAnalysisMLP} must be multiplied by the number of output variables of the model in order to obtain an approximate idea of the computational time they require.

\section{Conclusions} \label{sec:conclusions}

The \pkg{NeuralSens} package provides functions to extract information from a fitted feed-forward MLP neural network in \proglang{R}. These functions can be used to obtain the partial derivatives of the neural network response variables with regard to the input variables and to generate plots to obtain different information on the network using these partial derivatives. Methods are available for the next CRAN packages: \class{nn} (\pkg{neuralnet}), \class{nnet} (\pkg{nnet}), \class{mlp} (\pkg{RSNNS}), \class{H2ORegressionModel} and \class{H2OMultinomialModel} (\pkg{h2o}), \class{list} (\pkg{neural}), \class{nnetar} (\pkg{forecast}) and \class{train} (\pkg{caret}) (only if the object inherits the class attribute from another package). An additional method for class \class{numeric} is available to use with the basic information of the model (weights, structure and activation functions of the neurons).

The main objective of the package is to help the user to understand how the neural network uses the inputs to predict the output. This information may be useful for simplifying the neural network structure by eliminating the inputs which have no effect on the output. It could also provide a deeper understanding on the problem and the relationship among variables. \pkg{NeuralSens} is another tool among several other methods for exploratory data analysis and model evaluation, and it can be used with other packages (\cite{beck_neuralnettools_2018}, \cite{pdp}) to obtain more information on the neural network model. Nevertheless, it must be noted that sensitivity analysis using partial derivatives provides information about variable relationships such as pdp or ICE plots significantly faster. Moreover, it also provides variable importance measures like Garson's or Olden's methods and these measures are independent of the neural structure and training conditions of the model as long as it predicts the output with enough precision. 

Improving the information given by these methods will have value for exploratory data-analysis and characterization of relationships among variables. Future versions of the package may include additional functionalities as:
\begin{itemize}
	\item Parallelizing the sensitivity calculations when workers registered to work in parallel are detected.
	\item Calculating the sensitivities of the output variables with regard to the output of hidden neurons, in order to obtain the importance of each hidden neuron and helping to select the optimal network structure.
	\item Calculating the sensitivities of other neural network models such as Probabilistic Radial Basis Function Network (PRBFN) or Recurrent Neural Network (RNN).
	\item Calculating the second order partial derivatives of an MLP model to analyze the effect of interactions between two input variables.
	\item Develop a statistic to determine if the relationship between the output and input is linear.
\end{itemize}

\bibliography{jss4000}
\newpage

\begin{appendix}

\section[Examples of `SensAnalysisMLP' `numeric' method]{Examples of \fct{SensAnalysisMLP} \class{numeric} method} \label{app:ext_pack_func_impl}

\begin{leftbar}
The \proglang{R} packages \pkg{monmlp} \citep{monmlp} and \pkg{tensorflow} \citep{tensorflow} is used in this section to illustrate how the \class{numeric} method of \fct{SensAnalysisMLP} function can be applied to new models. The \code{simdata} dataset described in section \ref{subsec:Sensitivity} is used to train the MLP models in this appendix. 
\begin{CodeChunk}
\begin{CodeInput}
R> library("monmlp")
R> set.seed(150)
R> monmlp_model <- monmlp.fit(x = data.matrix(simdata[,1:3]),
+    y = data.matrix(simdata[,4]), hidden1 = 5, iter.max = 250,
+    silent = TRUE)
\end{CodeInput}
\end{CodeChunk}
The weights of the model are extracted and ordered as described in (\ref{eqn:neuralweightexample}):
\begin{CodeChunk}
\begin{CodeInput}
R> W1 <- rbind(monmlp_model[[1]]$W1[4,], monmlp_model[[1]]$W1[1:3,])
R> W2 <- c(monmlp_model[[1]]$W2[6,], monmlp_model[[1]]$W2[1:5,])
R> wts <- c(as.vector(W1), as.vector(W2))
\end{CodeInput}
\end{CodeChunk}
The activation functions of the hidden and output layers and their derivatives are provided in the \class{Th}, \class{To}, \class{Th.prime} and \class{To.prime} attribute of the model respectively. However, the derivative functions does not meet the condition of returning a square \class{matrix} so it must be modified before using \fct{SensAnalysisMLP}:
\begin{CodeChunk}
\begin{CodeInput}
R> Actfunc <- c("linear", attr(monmlp_model, "Th"), 
+    attr(monmlp_model, "To"))
R> Deractfunc <- c("linear",
+    function(v) {diag(attr(monmlp_model, "Th.prime")(v))},
+    function(v) {diag(attr(monmlp_model, "To.prime")(v))})
\end{CodeInput}
\end{CodeChunk}
The last information that must be passed to \fct{SensAnalysisMLP} is the neural structure:
\begin{CodeChunk}
\begin{CodeInput}
R> mlpstruct <- c(3,5,1)
\end{CodeInput}
\end{CodeChunk}
Since \fct{monmlp.fit} automatically scales the variables when training the model, the input variables must be scaled before calculating the sensitivities. 
\begin{CodeChunk}
\begin{CodeInput}
R> x <- data.matrix(simdata[,1:3])
R> x.center <- attr(monmlp_model, "x.center")
R> x.scale <- attr(monmlp_model, "x.scale")
R> x <- sweep(x, 2, x.center, "-")
R> x <- sweep(x, 2, x.scale, "/")
R> x <- cbind(data.frame(x),Y = simdata[,4])
\end{CodeInput}
\end{CodeChunk}
Once all the information has been prepared, the \class{numeric} method of \fct{SensAnalysisMLP} can be used to perform a sensitivity analysis of the model:
\begin{CodeChunk}
\begin{CodeInput}
R> sens_monmlp <- SensAnalysisMLP(wts, trData = x, mlpstr = mlpstruct,
+    coefnames = c("X1", "X2", "X3"), output_name = "Y",
+    actfunc = Actfunc, deractfunc = Deractfunc, plot = FALSE)
R> summary(sens_monmlp)
\end{CodeInput}
\begin{CodeOutput}
Sensitivity analysis of 3-5-1 MLP network.

Sensitivity measures of each output:
$Y
            mean         std  meanSensSQ
X1 -1.741261e-02 1.972087796 1.971671603
X2 -4.828116e-01 0.010075281 0.482916667
X3 -7.807139e-05 0.005014291 0.005013646
\end{CodeOutput}
\end{CodeChunk}
\fct{summary} method shows the same relationships between input variables \code{X1}, \code{X2} and \code{X3} and output variable \code{Y} as in sections \ref{subsec:Sensitivity} and \ref{subsec:Vis_Sensitivity}.

Sensitivity analysis of a \pkg{tensorflow} MLP model can be performed extracting analogous information as in the previous example. As the popularity of this package is growing rapidly,  a specific guide on how to extract the information seems necessary.

The following code is used to load the \pkg{tensorflow} and \pkg{keras} \citep{keras} libraries and to train a MLP model with two hidden layers:
\begin{CodeChunk}
\begin{CodeInput}
R> library("tensorflow")
R> library("keras")
R> keras_model <- keras_model_sequential() %>%
+    layer_dense(units = 16, activation = "relu", input_shape = 3) %>%
+    layer_dense(units = 8, activation = "relu") %>%
+    layer_dense(units = 1) %>%
+    compile(loss = "mse", optimizer = optimizer_rmsprop(),
+        metrics = list("mean_absolute_error"))
R> history <- keras_model %>% fit(data.matrix(simdata[,1:3]), 
+    array(simdata[,4]), epochs = 500, verbose = 0)
\end{CodeInput}
\end{CodeChunk}
Now that the model is trained, the weights and neural structure of the model can be obtained using the \fct{get\_weights} function:
\begin{CodeChunk}
\begin{CodeInput}
R> model_weights <- get_weights(keras_model)
R> wts <- c()
R> neural_struct <- c(nrow(model_weights[[1]]))
R> for (i in seq(2,length(model_weights),2)) {
+    neural_struct <- c(neural_struct,dim(model_weights[[i]]))
+    lyr_wgts <- rbind(model_weights[[i]],model_weights[[i-1]])
+    wts <- c(wts,unname(do.call(c, as.data.frame(lyr_wgts))))
+  }
\end{CodeInput}
\end{CodeChunk}
Since all the activation functions are already implemented in \pkg{NeuralSens}, they can be defined as a \class{character} \class{vector}.
\begin{CodeChunk}
\begin{CodeInput}
R> actfunc <- c("linear","ReLU","ReLU","linear")
\end{CodeInput}
\end{CodeChunk}
The \fct{SensAnalysisMLP} function can already be used with all the information obtained. 
\begin{CodeChunk}
\begin{CodeInput}
R> sens_keras <- SensAnalysisMLP(wts, trData = simdata, 
+    mlpstr = neural_struct, coefnames = names(simdata)[1:3],
+    output_name = names(simdata)[4], actfunc = actfunc, plot = FALSE)
R> summary(sens_keras)
\end{CodeInput}
\begin{CodeOutput}
Sensitivity analysis of 3-16-8-1 MLP network.

Sensitivity measures of each output:
$Y
           mean        std meanSensSQ
X1 -0.019691112 2.00377498 2.00337075
X2 -0.513628449 0.09752247 0.52280021
X3 -0.004615191 0.04967356 0.04987513
\end{CodeOutput}
\end{CodeChunk}

Again, the \fct{summary} method shows the same relationships between input variables \code{X1}, \code{X2} and \code{X3} and output variable \code{Y} as in sections \ref{subsec:Sensitivity} and \ref{subsec:Vis_Sensitivity}.
\end{leftbar}
\end{appendix}


\end{document}